\documentclass[twocolumn, switch]{article} 

\usepackage{algorithm,algorithmic}

\usepackage{bm}
\usepackage{subfig}
\usepackage{xurl}

\usepackage{preprint}

\usepackage{amsmath, amsthm, amssymb, amsfonts}

\usepackage[numbers,square]{natbib}
\usepackage[utf8]{inputenc}	
\usepackage[T1]{fontenc}	
\usepackage{xcolor}		
\usepackage[colorlinks = true,
            linkcolor = purple,
            urlcolor  = blue,
            citecolor = cyan,
            anchorcolor = black]{hyperref}	
\usepackage{booktabs} 		
\usepackage{nicefrac}		
\usepackage{microtype}		
\usepackage{lineno}		
\usepackage{float}			

\usepackage{lipsum}		

\usepackage{newfloat}
\DeclareFloatingEnvironment[name={Supplementary Figure}]{suppfigure}
\usepackage{sidecap}
\sidecaptionvpos{figure}{c}

\usepackage{titlesec}
\titlespacing\section{0pt}{12pt plus 3pt minus 3pt}{1pt plus 1pt minus 1pt}
\titlespacing\subsection{0pt}{10pt plus 3pt minus 3pt}{1pt plus 1pt minus 1pt}
\titlespacing\subsubsection{0pt}{8pt plus 3pt minus 3pt}{1pt plus 1pt minus 1pt}

\newcommand{\comment}[1]{}

\title{Semi-supervised Variational Autoencoder for Regression: Application on Soft Sensors}

\usepackage{authblk}

\author[1,*]{Yilin Zhuang}
\author[2,*]{Zhuobin Zhou}
\author[3]{Burak Alakent}
\author[1,+]{Mehmet Mercang\"{o}z}

\affil[1]{Department of Chemical Engineering, Imperial College London}
\affil[2]{Department of Mathematics, Imperial College London}
\affil[3]{Department of Chemical Engineering, Bogazici University}

\affil[*]{indicates equal contribution.}
\affil[+]{Correspondence: m.mercangoz@imperial.ac.uk}

\begin{document}

\twocolumn[ 
  \begin{@twocolumnfalse} 
  
\maketitle

\begin{abstract}
We present the development of a semi-supervised regression method using variational autoencoders (VAE), which is customized for use in soft sensing applications. We motivate the use of semi-supervised learning considering the fact that process quality variables are not collected at the same frequency as other process variables leading to many unlabelled records in operational datasets. These unlabelled records are not possible to use for training quality variable predictions based on supervised learning methods. Use of VAEs for unsupervised learning is well established and recently they were used for regression applications based on variational inference procedures. We extend this approach of supervised VAEs for regression (SVAER) to make it learn from unlabelled data leading to semi-supervised VAEs for regression (SSVAER), then we make further modifications to their architecture using additional regularization components to make SSVAER well suited for learning from both labelled and unlabelled process data. The probabilistic regressor resulting from the variational approach makes it possible to estimate the variance of the predictions simultaneously, which provides an uncertainty quantification along with the generated predictions. We provide an extensive comparative study of SSVAER with other publicly available semi-supervised and supervised learning methods on two benchmark problems using fixed-size datasets, where we vary the percentage of labelled data available for training. In these experiments, SSVAER achieves the lowest test errors in 11 of the 20 studied cases, compared to other methods where the second best gets 4 lowest test errors out of the 20.
\end{abstract}
\vspace{0.35cm}

  \end{@twocolumnfalse} 
] 

\section{Introduction}
\label{sec:introduction}
Advances in measurement technologies, increasing plant digitalization efforts, and the development of alternative data collection means, such as Internet of Things, are leading to more operational data being collected from industrial processes. However, it is still the case that some important quantities are hard to measure online due to challenges including delays in measurement, high cost of in-line sensors, and the difficulty of installing and  maintaining the in-line measurement setups. In response to these challenges, soft sensors have emerged as powerful tools for obtaining estimates of these hard-to-measure but important variables - commonly related to the quality of the production process and hence often referred to as quality or response variables - using variables that are readily available in the existing process - commonly referred to as process variables \cite{8941265, ZHUANG2022103747}.

Soft sensing methods can be generally categorized into first-principle and data-driven approaches. First principle-based soft sensors have the promise of extensive extrapolation capabilities as they capture the underlying physical phenomena, but in practice they tend to  suffer from several important drawbacks: complex processes can be affected by factors not included in the first-principal models used, the process operating conditions can exceed the empiricial correlations used in the first principle models, and configuring the first-principle models requires extensive subject matter expertise. Therefore, with increasing availability of data collected from industrial processes, data-driven or hybrid data-driven and first-principle models have become an attractive alternative to soft sensors using only first-principle models.

Data-driven soft sensor development commonly leverages linear methods such as principal component regression \cite{PCA1, PCA2}, partial least squares \cite{PLS1, PLS2}, and support vector regression \cite{SVR1}. More recently, artificial neural networks (ANNs) are utilized to extract nonlinear features and to model the underlying mechanisms of process behavior. Among various different neural network architectures, the Variational Autoencoder (VAE) and its different variants have attracted considerable attention for soft sensor development due to their capability of extracting complex distributions from process data\cite{VAE1, VAE2, VAE3}.

In process industries, process variables, such as temperature, flow rate and pressure, are usually measured at a high frequency, within seconds to minutes, while quality variables, such as composition are usually measured at a slower rate, within minutes to hours. This practice yields a larger number of unlabelled data points, comprising only process variable measurements, compared to labelled data points, consisting of both process and quality variable measurements. Within the perspective of traditional supervised learning methods, these unlabelled process records cannot be utilized to train a model. To address this problem, researchers have shown an increased interest in developing semi-supervised data-driven soft sensors to improve the estimation accuracy of quality variables by utilizing the information contained in the unlabelled portions of the collected data. The VAE was first proposed as a generative neural network for image data \cite{Kingma_VAE}, it approximates the evidence lower bound (ELBO) as a tractable lower bound to the marginal likelihood of the data and the approximate inference distribution. Several VAE regression structures have been designed to perform a regression task based on inputs \cite{VAE3, VAER1, VAER2}, however, most of these designs set the regression module as a separate term in the loss function, i.e., the loss function consists of an ELBO term and a regression term. This segregation can lead to a problem that a good ELBO value does not always lead to an improvement in the inference result\cite{info_vae}. As a potential solution to this problem, it was suggested to incorporate labelled information in the ELBO \cite{shot_vae}, and a unified supervised VAE for regression (SVAER) \cite{VAER_zhao} was proposed to obtain a latent representation of the inputs that conditioned on the labelled information. The presented results show improvements in quality variable prediction compared to several other neural architectures. The SVAER was applied on brain image analysis where the quality variable is the age of the subject associated with the image. The conditioned correlation enables the generation of images from the quality variable. Although not pointed out by the authors, one should notice that the feasibility of conditioning the latent space on the quality variable relies on an implicit constraint: the mapping of the quality variable to the latent space is injective. Therefore, the SVAER could suffer for performing regression on process data because the function of mapping the quality variable to the latent space is usually one-to-many. In addition, the regression model proposed in \cite{VAER_zhao} was designed for supervised learning, and additional modifications are needed to extend its structures to be well suited for semi-supervised learning, which would be required to deal with the case of unlabelled entries for quality variables.

In this study, we extend the work in \cite{VAER_zhao}, and construct a semi-supervised VAE for regression (SSVAER) that can utilize unlabelled process data to improve the quality variable prediction performance. Furthermore, we introduce an additional regressor block to account for the non-injective mapping between the quality variable and the latent space. It has been shown in dynamical systems that predicting further steps and taking those predictions into account in the loss function can benefit the accuracy and the stability of iterative predictions \cite{solver_in_loop}, with that inspiration in mind, the newly added regressor block aims to infer a pseudo change in the quality variable that best matches the latent representation in the next step. The main contributions of the presented work are as follows:
\begin{itemize}
    \item An entropy minimising term is introduced to the loss function and we derived the ELBO of the new SSVAER. A reconstruction based regularization term is added to improve the robustness of the model, and the SSVAER can inherit the advantages of SVAER to learn the latent representation that is correlated to the quality variables.
    \item We demonstrated the SSVAER can generally outperform SVAER on the two selected datasets, and can perform better than several other publicly available semi-supervised learning methods.
    \item We illustrated the SSVAER can also extract a structured latent representation and discussed its potential applications. We also showed the SSVAER can provide a reasonable estimation on the variance of the predicted quality variable.
\end{itemize}

The rest of the paper is structured as follows: in Sec. \ref{sec:pre}, we introduce the background on the VAE and the SVAER, the structure of the SSVAER is presented in Sec. \ref{sec:ssvaer}, two benchmark industrial processes are introduced in Sec. \ref{sec:case}, the results and discussion are presented in Sec. \ref{sec:result}. The implementation of the SSVAER is available at: \url{https://github.com/tonyzyl/Semisupervised-VAE-for-Regression-Application-on-Soft-Sensor}

\section{Preliminaries \label{sec:pre}}

\subsection{Variational Autoencoder}

The original VAE \cite{Kingma_VAE} consists of an encoding and a decoding network that are made of fully connected (FC) layers. The encoder $q(\bm{z}|\bm{x})$ maps the input, $\bm{x}\in \mathbb{R}^{n}$, to its latent representation, $\bm{z}\in \mathbb{R}^{d}$, where $q$ denotes the encoding network. The distributions of latent values are generally assumed as normal, and usually the standard Gaussian distribution is chosen, i.e., $p(z) = \mathcal{N}\left(\bm{z}|\bm{0},\textbf{I}\right)$. Similarly, the decoding network can be expressed as $p(\bm{x}|\bm{z})$, where $p$ denotes the decoding network, $\bm{z}$ is resampled from the latent distribution given by the encoding network by using the reparameterization trick \cite{elbo1}. The objective of the generative network is to find the distribution, $p(\bm{x})$, that best fits the inputs, however, the computation of the marginal likelihood, $p(\bm{x})=\int p(\bm{x}|\bm{z})p(\bm{z}) d\bm{z}$, is intractable, and an amortized inference distribution $q(\bm{z}|\bm{x})$ is introduced to jointly optimize the ELBO of the log marginal likelihood \cite{elbo1}:
\begin{align}
\begin{split}
    \log p(\bm{x})  =\, & \mathbb{E}_{q(\bm{z}|\bm{x})}\left[\log p(\bm{x},\bm{z})-\log q(\bm{z}|\bm{x})\right]\\&+D_{KL}\left(q(\bm{z}|\bm{x})\parallel p(\bm{z}|\bm{x})\right)\\
    \geq &\, \mathbb{E}_{q(\bm{z}|\bm{x})}\left[\log p(\bm{x},\bm{z})-\log q(\bm{z}|\bm{x})\right]\\
    (ELBO, \mathcal{L}) =\, &\mathbb{E}_{q(\bm{z}|\bm{x})}\left[\log p(\bm{x},\bm{z})\right]-D_{KL}\left(q(\bm{z}|\bm{x})\parallel p(\bm{z})\right)
\end{split}
\label{eq:vae_elbo}
\end{align}
The $D_{KL}(\cdot || \cdot )$ in Eq. \eqref{eq:vae_elbo} denotes the Kullback-Leibler (KL) divergence, it quantifies the differences between two distributions, and by Jensen's inequality, the KL divergence is always greater or equal to zero. The ELBO$(\mathcal{L})$ consists of the expected conditional log likelihood and the KL divergence between the variational distribution and the prior. In VAE, the encoding and decoding network, $q_\phi(\bm{z}|\bm{x})$ and $p_\theta(\bm{x}|\bm{z})$ are parameterized by $\phi$ and $\theta$, denoting the parameters of the neural networks. The stochastic gradient variational Bayes (SGVB) estimator \cite{Kingma_VAE} is used to compute the ELBO in the training process, and the weights of the neural networks are optimized by back propagation.

\subsection{Supervised VAE for regression}
\begin{figure}[!htb]
\centerline{\includegraphics[width=\columnwidth]{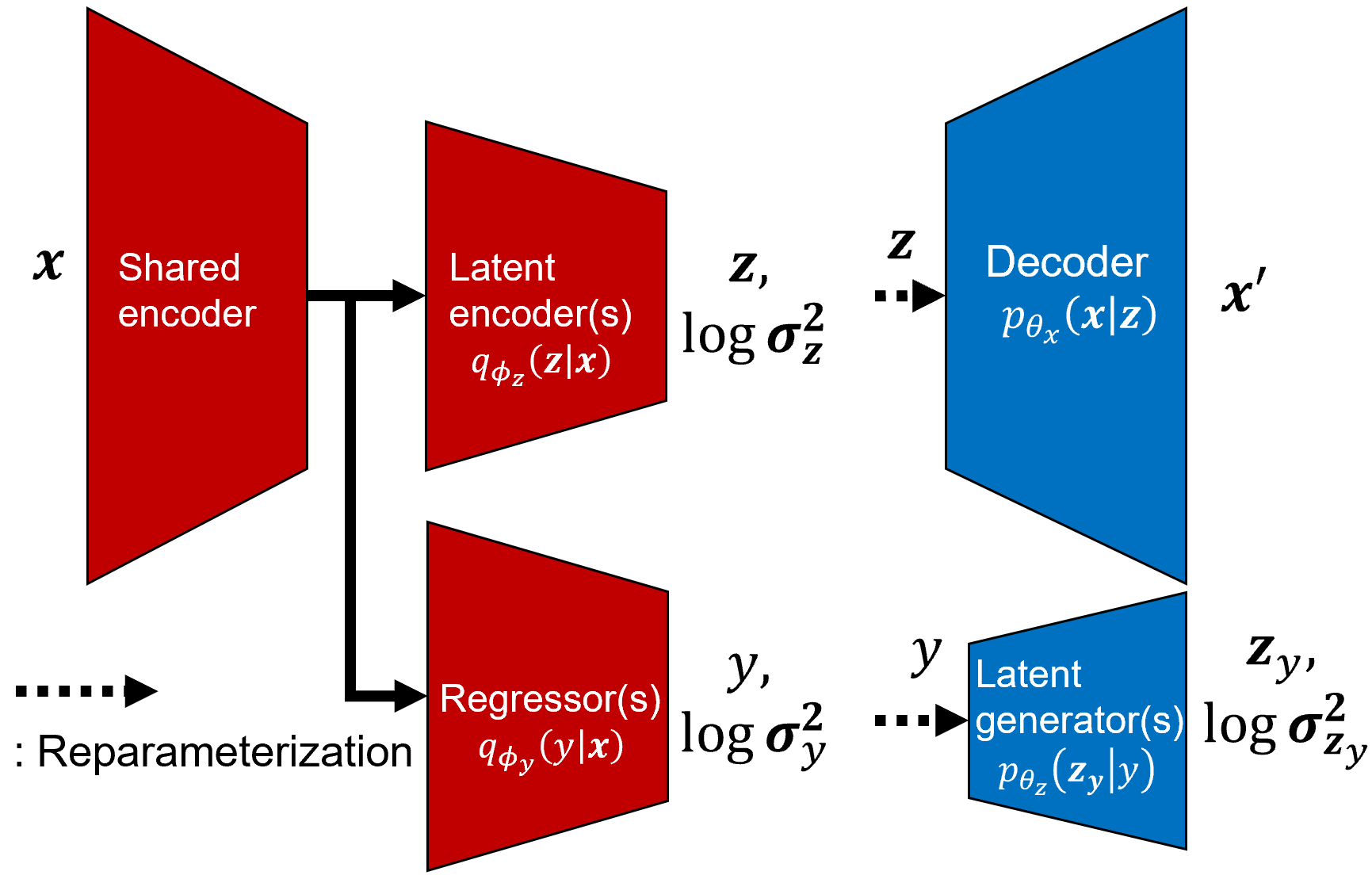}}
\caption{VAE-based regression model \cite{VAER_zhao}.}
\label{fig:svaer}
\end{figure}

Compared to the VAE, the SVAER\cite{VAER_zhao} has a probabilistic regressor to estimate the quality variable, $y$, $q_{\phi_y}(y|\bm{x}) \sim \mathcal{N}\left(y|f(\bm{x};\phi_y),g(\bm{x};\phi_y)^2\right)$, where $f$, $g$ denote two FC neural networks that are parameterized by $\phi_y$. The schematic of the SVAER is shown in Fig. \ref{fig:svaer}. The extracted distribution from the dataset is regularized during the training by a latent generator, $p_{\theta_z}(\bm{z}|y)\sim \mathcal{N}\left(\bm{z}|\bm{W}^Ty,\sigma^2\textbf{I})\right)$, $\bm{W}^T\bm{W}=1$, where $\sigma^2$ denotes the variance of the quality variable estimated by the probabilistic regressor. The latent generator maps the regressed quality variable to the corresponding latent values, and their differences are quantified by the KL divergence. With this setup, it can be seen that the latent values are now varying by the given quality variable, while leaving all other parameters invariant, i.e., the structure disentangles the quality variable from the latent space \cite{disentangle_higg}. Similar to VAE, the training objective is to extract a distribution that can best describe the inputs, and the expression of the log marginal likelihood can be written as:
\begin{align}
    \begin{split}
        \log p(\bm{x}) = &\int_{\bm{z}}\int_{y} q(\bm{z}, y|\bm{x})\log p(\bm{x})\,d\bm{z} \,dy \\
        = & D_{KL}\left(q_{\phi_y}((\bm{z}, y|\bm{x})\parallel p((\bm{z}, y|\bm{x})\right)+\mathcal{L}(\bm{x})
    \end{split}
\end{align}
Where the ELBO of the SVAER can be expressed as:
\begin{align}
    \begin{split}
        \mathcal{L}(\bm{x}) = & \int_{\bm{z}}\int_{y} q(\bm{z}, y|\bm{x})\log \left(\frac{p(\bm{z},y,\bm{x})}{q(\bm{z}, y|\bm{x})}\right)\,d\bm{z} \,dy \\
        = & \int_{\bm{z}}\int_{y} q(\bm{z}, y|\bm{x})\log \left(\frac{\overbrace{p(\bm{x}|\bm{z},y)}^{\textcircled{1}}\overbrace{p(\bm{z}|y)}^{\textcircled{2}}\overbrace{p(y)}^{\textcircled{3}}}{q(\bm{z}, y|\bm{x})}\right)\,d\bm{z} \,dy\\
    \end{split}
    \label{eq:svaer_elbo}
\end{align}
The authors assume $q(\bm{z}, y|\bm{x}) = q(\bm{z}|\bm{x})q(y|\bm{x})$, i.e. the latent variables and the quality variable generated from the same inputs are independent.  Another assumption made is that $p(\bm{x}|\bm{z},y)=p(\bm{x}|\bm{z})$, i.e., the quality variable is disentangled from the learnt latent representation, the three annotated terms in Eq. \eqref{eq:svaer_elbo} can then be separated as:
\begin{align}
\begin{split}
    \text{Rec. loss} &=  \int_{\bm{z}}\int_{y} q(y|\bm{x}) \, dy\, q(\bm{z}|\bm{x})\log p(\bm{x}|\bm{z})\, d\bm{z}\\
    &= \int_{\bm{z}} q(\bm{z}|\bm{x})\log p(\bm{x}|\bm{z})\,d\bm{z} \\
     &=  \mathbb{E}_{q_{\phi_z}(\bm{z}|\bm{x})}\log p_{\theta_x}(\bm{x}|\bm{z})\label{eq:svaer_rec}
\end{split}\\
\begin{split}
    \text{KL loss} &=  \int_{\bm{z}}\int_{y} q(\bm{z}, y|\bm{x})\log \left(\frac{p(\bm{z}|y)}{q(\bm{z}|\bm{x})}\right)\,d\bm{z} \,dy \\
     &=  -\mathbb{E}_{q_{\phi_y}(y|\bm{x})}\left[D_{KL}\left(q_{\phi_z}(\bm{z}|\bm{x})\parallel p_{\theta_z}(\bm{z}|y)\right)\right]\label{eq:svaer_kl}
\end{split}\\
\begin{split}
    \text{Label loss} &=  \int_{y} q(y|\bm{x})\log\left(\frac{p(y)}{q(y|\bm{x})}\right)\,dy \\
     &=  -D_{KL}\left(q_{\phi_y}(y|\bm{x})\parallel p(y)\right)\label{eq:svaer_label}
\end{split}
\end{align}

With the formulations given in Eq. \eqref{eq:svaer_rec}-\eqref{eq:svaer_label}, the regression on the quality variable can be jointly trained with the two regularization terms, the reconstruction and the KL divergence between the prior generated from the regressed quality variable, $p_{\theta_z}(\bm{z}_y|y)$, and the approximate prior, $q_{\phi_y}(y|\bm{x})$.

\section{Semi-supervised VAE for regression}
\label{sec:ssvaer}

\begin{figure*}[!htb]
    \centerline{\includegraphics[width=2\columnwidth]{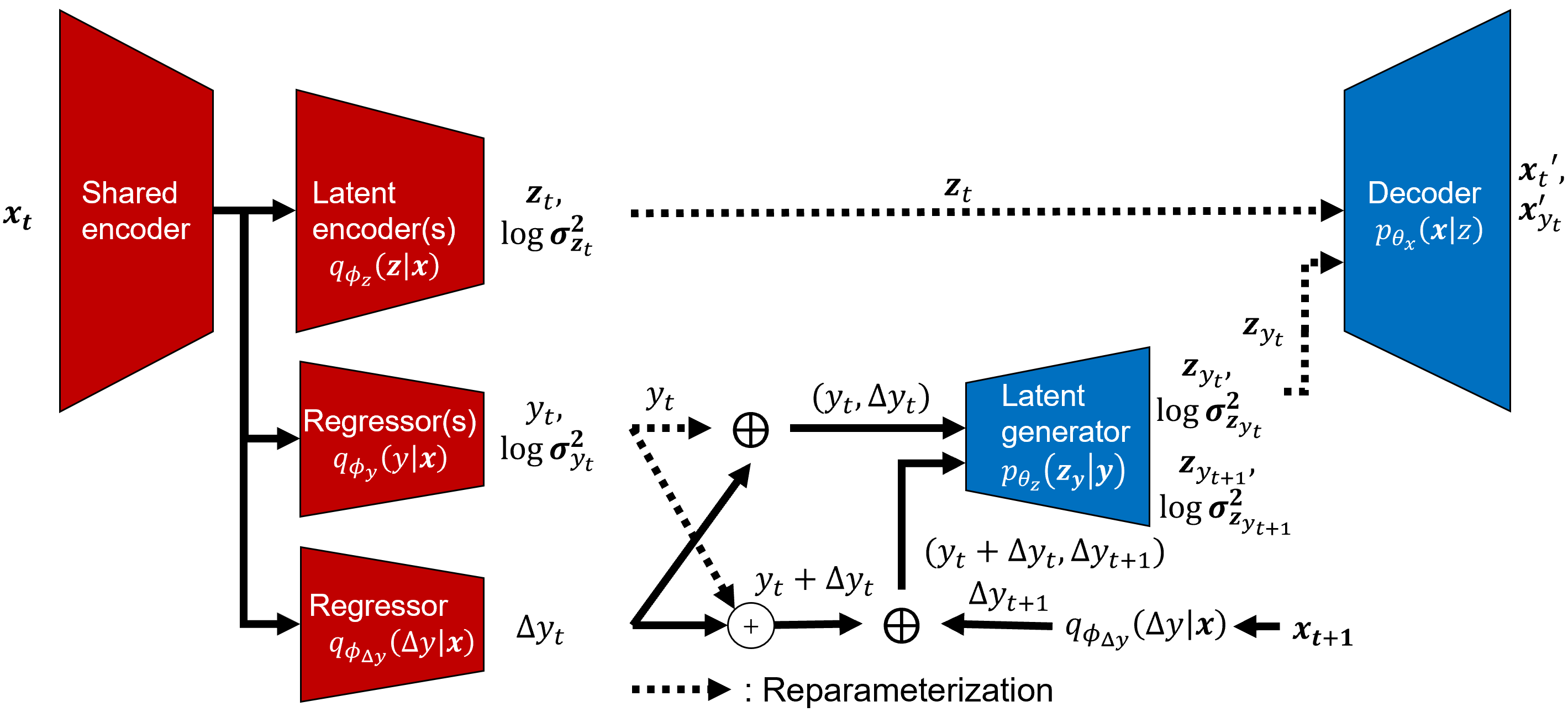}}
    \caption{The schematic of SSVAER, "$\oplus$" denotes the concatenation of inputs, "$\textcircled{+}$" denotes the summation of inputs. Neural networks colored in red denotes the inference network, whereas blocks colored in blue denotes the generative network.}
    \label{fig:ssvaer_scheme}
\end{figure*}

The schematic of the full SSVAER model is shown in Fig. \ref{fig:ssvaer_scheme}. In this section, we will firstly extend the SVAER to make it learn from unlabelled data, then we will introduce the additional modifications that make SSVAER well suited for learning from process data. In the rest of this paper, we will use subscripts $l$ and $u$, to denote labelled and unlabelled data, respectively. 

\subsection{Extension on semi-supervised learning}

From inspection of terms of the loss functions of the SVAER, Eq. \eqref{eq:svaer_rec}-\eqref{eq:svaer_label}, it can be seen that the label information, $p(y)$, only contributes in the label loss term. One common approach is to replace the label loss term with the entropy of the quality variable's variational distribution, $\mathbb{H}\left[q_{\phi_y}(y|\bm{x}_u)\right]$ \cite{Kingma_SVAE, semi_entropy}. The ELBO for the labelled and unlabelled entry can be formulated as:
\begin{align}
    \begin{split}
        \mathcal{L}_l(\bm{x}) =& \mathbb{E}_{q_{\phi_z}(\bm{z}|\bm{x})}\log p_{\theta_x}(\bm{x}|\bm{z})-\mathbb{E}_{q_{\phi_y}(y|\bm{x})} [D_{KL}(q_{\phi_z}(\bm{z}|\bm{x})\parallel \\ & p_{\theta_z}(\bm{z}|y) )] -D_{KL}\left(q_{\phi_y}(y|\bm{x})\parallel p(y)\right)
        \label{eq:label_loss}
    \end{split}
\end{align}
\begin{align}
    \begin{split}
        \mathcal{L}_u(\bm{x}) =& \mathbb{E}_{q_{\phi_z}(\bm{z}|\bm{x})}\log p_{\theta_x}(\bm{x}|\bm{z})-\mathbb{E}_{q_{\phi_y}(y|\bm{x})}\left[D_{KL}\left(q_{\phi_z}(\bm{z}|\bm{x})\right.\right.\parallel \\
        &  p_{\theta_z(\bm{z}|y)})] +\mathbb{H}\left[q_{\phi_y}(y|\bm{x})\right]
        \label{eq:unlabel_loss}
    \end{split}
\end{align}

The subscripts $l$ and $u$, are dropped in Eqs. \eqref{eq:label_loss}, \eqref{eq:unlabel_loss} for a clearer expression. For computing the entropy $\mathbb{H}\left[q_{\phi_y}(y|\bm{x}_u)\right]$, we will assume the learnt approximate prior, $q_{\phi_y}(y|\bm{x}_u)$, follows a normal distribution, and it can be computed directly from:
\begin{align}
    \mathbb{H}\left[q_{\phi_y}(y|\bm{x}_u)\right]=-\mathbb{E}\left[\mathcal{N}\left(y|f(\bm{x}_u;\phi_y),g(\bm{x}_u;\phi_y)^2\right)\right]
\end{align}

In previous models, the reconstruction loss does not directly contribute to the training of the latent generator. In order to prevent overfitting due to diminishing KL loss during the training, we propose an additional reconstruction regularization term,
\begin{align}
    \begin{split}
        \mathcal{L}_u'(\bm{x}) &= \mathcal{L}_u(\bm{x}) + \mathbb{E}_{q_{\phi_z}(z|x)}log(p_{\theta_x}(x|z_{y}))
    \end{split}
\end{align}
where z is generated from $q_{\phi_z}(z|x)$ and $z_{y}$ is generated from $p_{\theta_z}(z_{y}|y)$ as shown in Fig. \ref{fig:ssvaer_scheme}. The loss generated from this term will skip the decoder when updating weights by back propagation, so that this term is directly enacted on the inference network of the quality variable.

\subsection{Extension on pseudo variation estimation}
\label{sec:pseudo}

In order to effectively disentangle the scalar quality variable from the latent representation, at least one more degree of freedom (DOF) should be added to account for the variations of the quality variable. To come up with this new scalar, a new FC regressor is added, and it is selected to have the same number of neurons as the quality variable regressor. 

Although this new DOF can be any arbitrary value, i.e., discard the contribution of this term in the loss function and let the SSVAER infer this value in a self-supervised manner, the new DOF acts as an input to the inference network therefore it should be assigned as a certain measurable or physical property if possible. Since most process records are time-series data, it is sensible to select the new DOF to be the change of the quality variable from the previous time step. Denoting this new DOF as $\Delta y$, intuitively, we would like to have $y_{t+1}=y_t+\Delta y_t$, where the subscript $t$, $t+1$, represents the current, and the following time step, respectively. The latent generator can now be written as $p_{\theta_z}(\bm{z}|\bm{y})$, where $\bm{y}$ denotes the concatenation of $y$, $\Delta y$. 

For semi-supervised learning, the labels are not always available, thus, it is impractical to directly include the residual of $y_{t+1}-y_t-\Delta y_t$ in the loss function. The latent representations of process inputs can be utilized to circumvent this problem, because they are always available for all time steps, and the new latent generator can map $y$ and $\Delta y$ to the latent representation. In SVAER, the latent representation is linearly correlated to the quality variable, however in SSVAER, nonlinear correlation is preferred to capture more information on the dynamical system. Under this construction, the newly introduced DOF, $\Delta y$, can be regularized by comparing the differences between the latent representation generated from $(y_t+\Delta y_t,\Delta y_{t+1})$, and the actual latent representation encoded from the input of the following input, $x_{t+1}$. The difference can then be quantified by the KL divergence, denoted as the pseudo variation loss (P.V. loss):
\begin{align}
\begin{split}
    &\text{P.V. loss} = \\
    &-D_{KL}\left(q_{\phi_z}(\bm{z}_{t+1}|\bm{x}_{t+1})\parallel p_{\theta_z}(\bm{z}_{t+1}|(y_t+\Delta y_t,\Delta y_{t+1}))\right)
    \label{eq:dy_kl}
\end{split}
\end{align}

Then the regressor for $\Delta y$ will try to infer the change in the quality variable that matches the input of the next time step. Since this $\Delta y$ does not relate to the actual variation in the quality variable, we will refer this term as pseudo variation. Note that the regressors for $y$ and $\Delta y$ are separated, thus, the information of the future time step is only required to train the SSVAER. The functioning of the regressor for quality variable will not be affected by other parts during the online inference, i.e., all the parts, other than the shared encoder and the quality variable regressor, serve as regularization terms to the label loss in the loss function.

\subsection{Structure and implementation of SSVAER}

As the structure of SSVAER shown in Fig. \ref{fig:ssvaer_scheme}, the inference part consists of a latent encoder, a quality variable regressor, and a pseudo variation regressor, that are represented as  $q_{\phi_z}(\bm{z}|\bm{x})\sim \mathcal{N}\left(\bm{z}|f(\bm{x};\phi_z),g(\bm{x};\phi_z)^2\right)$, $q_{\phi_y}(y|\bm{x}) \sim \mathcal{N}\left(y|f(\bm{x};\phi_y),g(\bm{x};\phi_y)^2\right)$, and $q_{\phi_{\Delta y}}(\Delta y|\bm{x}) = f(\bm{x};\phi_{\Delta y})$, respectively. The generative part consists of a latent generator and a decoder, that are represented as $p_{\theta_z}(\bm{z}_y|\bm{y})\sim\mathcal{N}\left(\bm{z}|f((y,\Delta y));\phi_{z_y}),g((y,\Delta y);\phi_{z_y})^2\right)$, and $p_{\theta_x}(\bm{x}|\bm{z})\sim\mathcal{N}\left(\bm{x}|f(\bm{z};\phi_x),g(\bm{x};\phi_x)^2\right)$, respectively. In practice, we select the latent generator to follow a three layer structure with size of \{2,2,$N_z$\}, where the number in the curly bracket denotes the number of neurons in each layer, and $N_z$ denotes the size of latent space. It follows the intuition of maintaining the DOF while allowing some interactions between $y$ and $\Delta y$.

With the added pseudo variation regressor, and ignoring the differentials of variables in the following expression for simplicity, the ELBO of SSVAER can then be formulated as:
\begin{align}
    \begin{split}
        &\mathcal{L}(\bm{x}) =  \int_{\bm{z}}\int_{y,\Delta y}q(\bm{z}, y,\Delta y|\bm{x})\log \left(\frac{p(\bm{z},y,\Delta y,\bm{x})}{q(\bm{z}, y,\Delta y|\bm{x})}\right)  \\
    \end{split}
    \label{eq:ssvaer_elbo}
\end{align}
and the natural logarithm can be expanded as:
\begin{align*}
    \log \left(\frac{\overbrace{p(\bm{x}|\bm{z},y,\Delta y)}^{\textcircled{1}}\overbrace{p(\bm{z}|y,\Delta y)}^{\textcircled{2}}\overbrace{p(y,\Delta y)}^{\textcircled{3}}}{q(\bm{z}, y,\Delta y|\bm{x})}\right)
\end{align*}

In this case, we extend the assumption in SVAER to $q(z,y,\Delta y|x ) = q(z|x)q(y|x)q(\Delta y |x)$ and further assume that the true distributions of $y$ and $\Delta y$ are independent, i.e., $p(y,\Delta y) = p(y)p(\Delta y)$. Rewriting Eq. \eqref{eq:svaer_rec}-\eqref{eq:svaer_label}, we obtain four new simplified losses:
\begin{align}
\begin{split}
     \text{Rec. loss}&=  \mathbb{E}_{q_{\phi_z}(\bm{z}|\bm{x})}\log p_{\theta_x}(\bm{x}|\bm{z})\label{eq:ssvaer_rec}
\end{split}\\
\begin{split}
     \text{KL loss} &=  -\mathbb{E}_{q_{\phi_y}(y,\Delta y|\bm{x})}\left[D_{KL}\left(q_{\phi_z}(\bm{z}|\bm{x})\parallel p_{\theta_z}(\bm{z}|y,\Delta y)\right)\right]\label{eq:ssvaer_kl}
\end{split}\\
\begin{split}
     \text{Label loss} &=  -D_{KL}\left(q_{\phi_y}(y|\bm{x})\parallel p(y)\right)\label{eq:ssvaer_label}
\end{split}\\
\begin{split}
    \text{$\Delta y$ loss}  &=  -D_{KL}\left(q_{\phi_{\Delta y}}(\Delta y|\bm{x})\parallel p(\Delta y)\right)\label{eq:ssvaer_dy}
\end{split}
\end{align}
As mentioned in Sec. \ref{sec:pseudo}, it is impractical to compute the the label loss for $\Delta y$ directly. Hence we replace the $\Delta y$ loss with Eq.\eqref{eq:dy_kl}. The new ELBO for the labelled and unlabelled entries can then be formulated as:
\begin{align}
    \begin{split}
        \mathcal{L}_l(\bm{x}) = \text{Rec.}+\text{KL}+\text{P.V.}-D_{KL}\left(q_{\phi_y}(y|\bm{x})\parallel p(y)\right)
        \label{eq:label_loss_new}
    \end{split}
\end{align}
\begin{align}
    \begin{split}
       \mathcal{L}_u(\bm{x})= \text{Rec.}+\text{KL}+\text{P.V.}+\mathbb{H}\left[q_{\phi_y}(y|\bm{x})\right]
        \label{eq:unlabel_loss_new}
    \end{split}
\end{align}
The overall loss function, $\mathcal{J}(\bm{x})$, can then be formulated as:
\begin{align}
    \mathcal{J}(\bm{x}_t,\bm{x}_{t+1}) =& \mathcal{L}_l(\bm{x}_t) + \mathcal{L}_u(\bm{x}_t,\bm{x}_{t+1})+ \\ & \mathbb{E}_{q_{\phi_z}(\bm{z}|\bm{x})}\log(p_{\theta_x}(x_{t}|z_{y}))
    \label{eq:ssvaer_loss_func}
\end{align}

Two consecutive records of inputs are needed in the forward path of training the SSVAER. It can be seen from Fig. \ref{fig:ssvaer_scheme}, regressors for the pseudo variation and the latent generator are called for both time steps, whereas all other parts are only called for the current time step. As the following time steps will be processed in the future training loops, it is sensible to let the loss to only back propagate through the pseudo variation regressor and the latent generator once, to reduce the interference caused by the next time step. Eq. \eqref{eq:ssvaer_loss_func} can then be calculated by the SGVB estimator, and the weights of the neural networks are updated by back propagation. An algorithm demonstrating the training procedure of the SSVAER is outlined in Algo. \ref{algo:1}.

\begin{algorithm}
\caption{Algorithm for SSVAER}\label{algo:1}
\begin{algorithmic}[1]
\renewcommand{\algorithmicrequire}{\textbf{Inference input:}}
\renewcommand{\algorithmicensure}{\textbf{Inference output:}}
\REQUIRE Process inputs at time step $t$, $\bm{x}_t$.
\ENSURE  Quality variable at time step $t$, $y_t$
\\ \textit{Initialization} : Generate the SSVAER model based on the number of neurons provided for each module, and initialized the set of weights $\{\phi\}$.
\\ \textit{Training loop}
\FOR {\textit{Epoch} $= 1$ to N}
\STATE Compute latent rep.: $\bm{z}_t$, $\log\bm{\sigma}^2_{z_t}\xleftarrow{q_{\phi_z}(\bm{z}|\bm{x})}\bm{x}_t$
\STATE Compute quality variable: $y_t$, $\log\bm{\sigma}^2_{y_t}\xleftarrow{q_{\phi_y}(y|\bm{x})}\bm{x}_t$
\STATE Compute pseudo variation for the two time steps: $\Delta y_t \xleftarrow{q_{\phi_{\Delta y}}(\Delta y|\bm{x})}\bm{x}_t$, $\Delta y_{t+1} \xleftarrow{q_{\phi_{\Delta y}}(\Delta y|\bm{x})}\bm{x}_{t+1}$
\STATE Generate latent rep. for time step $t$: \\$\bm{z}_{y_t}$, $\log\bm{\sigma}^2_{z_{y_t}}\xleftarrow{p_{\theta_z}(\bm{z}_y|\bm{y})}(y_t,\Delta y_t)$
\STATE Approximate latent rep. for time step $t+1$: \\$\bm{z}_{y_{t+1}}$, $\log\bm{\sigma}^2_{z_{y_{t+1}}}\xleftarrow{p_{\theta_z}(\bm{z}_y|\bm{y})}(y_t+\Delta y_t,\Delta y_{t+1})$
\STATE Compute loss function: $\mathcal{J}(\bm{x}_t,\bm{x}_{t+1}) = \mathcal{L}_l(\bm{x}_t) + \mathcal{L}_u'(\bm{x}_t,\bm{x}_{t+1})$
\STATE Weights back propagation: $\{\phi\}\leftarrow\mathcal{J}(\bm{x}_t,\bm{x}_{t+1})$
\ENDFOR
\\ \textit{Quality variable inference}
\STATE Compute quality variable: $y_t\xleftarrow{q_{\phi_y}(y|\bm{x})}\bm{x}_t$
\RETURN $y_t$ 
\end{algorithmic} 
\end{algorithm}

\section{Case studies}
\label{sec:case}

In this section, we will demonstrate the results of the SSVAER model on two publicly available benchmark datasets \cite{benchmark_data}, and to compare its performance with SVAER and several other methods that is either publicly available or there is existing publicly available implementation. These codes include, a layer-wise trainned semi-supervised autoencoder (SSAE) \cite{ssae}, and a semi-supervised Student’s t mixture model (SSSMM) \cite{SSSMM}. In addition, we add the supervised FC neural network (FCNN) into comparison, which is one of the basic blocks of all these networks. In order to make fair comparison between these methods, quality variable inference networks of these neural network methods are kept the same, i.e., number of layers, number of neurons of each layer, the activation function of each layer, and the weight initialization of each type of layers are the same. For the SSSMM, we select the number of mixing components to match the size of the latent space of neural network methods.

The two datasets are all fully labelled, and a certain portion of labels need to be removed to simulate semi-supervised learning scenarios. In this study, we would investigate the performance of each method with $1\%$, $2\%$, $5\%$, $10\%$, $14.2\%$, $20\%$, $25\%$, $33.3\%$, $50\%$, $100\%$ labelled entries in the training and validation datasets. For the supervised methods, only the labelled process records are used in the training. The labelled entries are selected following a regular pattern to simulate a common scenario that the sampling frequencies of process variables and quality variable are inconsistent, for instance, having $10\%$ labelled entries would correspond to a scenario that a quality variable is recorded for every 10 process records in a row. It should be noted that this construction would make the labelled information in each testing cases to be different from each other.

Adam optimizer \cite{adam} is used to calculate the gradient of neural network methods, batch size is selected to be 200, and the total epoch is selected to be 300. In the SSSMM, the maximum iteration is set to be 300. To train these neural networks, we use a varying learning rate between 0.01 and 0.0001, with a cosine annealing schedule \cite{sgdr} and 60 epochs are allocated for warming up. For the SSAE, each layer of the inference network is pretrained with a learning rate of 0.01 for 10 epochs, these are accounted as warming up and we subtract them from the total epochs and warm up epochs. In addition, the weights that obtain the least value from applying the loss function on the validation dataset will be saved for each method, and will be used to compute the prediction on test dataset.

\subsection{Debutanizer column}

\begin{figure}[!htb]
\centerline{\includegraphics[width=\columnwidth]{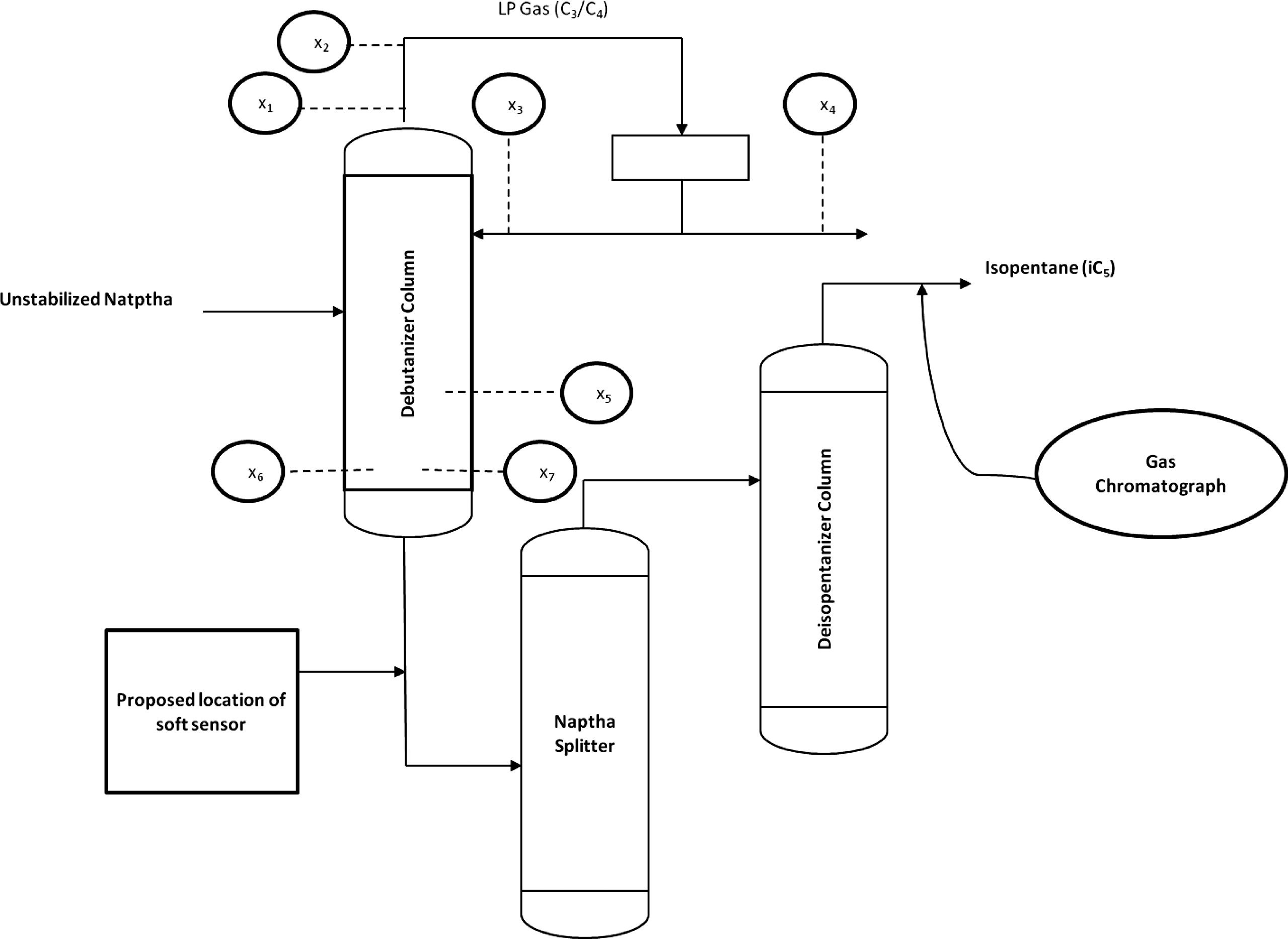}}
\caption{Schematic diagram of the debutanizer column, figured by \cite{debut_figure}.}
\label{fig:debut_scheme}
\end{figure}

The schematic of the debutanizer column is shown in Fig. \ref{fig:debut_scheme}, there is no online measurement for the butane (C4) content in the bottom product of the debutanizer column, thus, there is a measurement delay for the composition of C4. The process variables and the quality variable of the debutanizer column is shown in Tab. \ref{tab:debut_var}. 

\begin{table}[!htb]
\centering
\caption{Annotations for process variables of C4 column.}
\label{tab:debut_var}
\begin{tabular}{cc}
\toprule
Variables & Description   \\
\midrule
$x_1$ & Top temperature\\
$x_2$  & Top pressure\\
$x_3$  & Reflux flow\\
$x_4$  & Flow to next process\\
$x_5$  & 6th tray temperature\\
$x_6$  & Bottom temperature\\
$x_7$  & Bottom temperature\\
$y$ & Bottom C4 composition\\
\bottomrule
\end{tabular}
\end{table}

One common way of constructing the input to the model is to include the past records of process variables, e.g., for $x_1$, $ x_1(t)$, $x_1(t-5)$, $x_1(t-7)$, and $x_1(t-9)$ are included. Unlike the debutanizer data, no past records of the quality variable are included in the input, thus, no further assumptions are made. The input of the SRU problem can then be expressed as:

\begin{align}
\begin{split}
    \bm{x}(t) = &[ x_1(t),\dots, x_1(t-9),x_2(t),\dots,x_2(t-9), x_3(t),\dots,\\& x_3(t-9), x_4(t),\dots, x_4(t-9), x_5(t),\dots, x_5(t-9) ]
    \end{split}
    \label{eq:sru_input}
\end{align}

\begin{figure*}[t]
\centerline{\includegraphics[width=2\columnwidth]{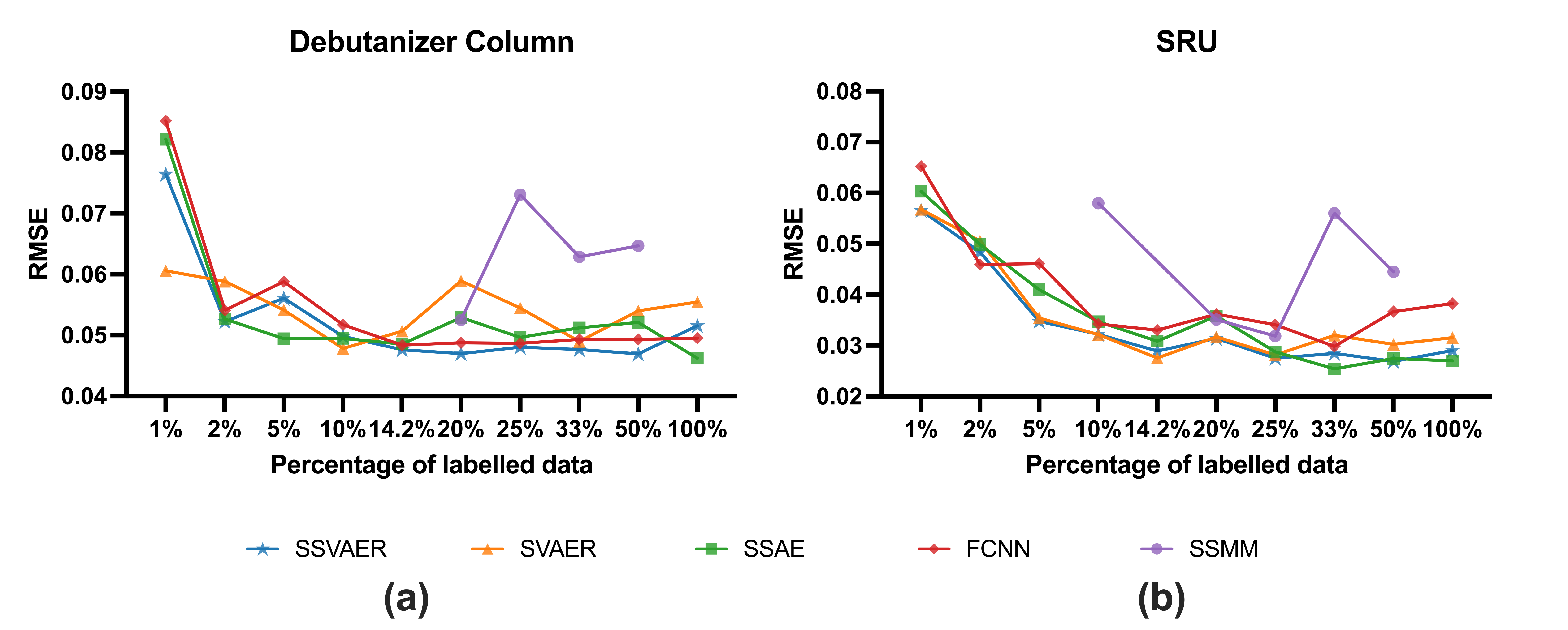}}
\caption{Variations of RMSEs against percentage of labelled entries on test datasets.}
\label{fig:label_trend}
\end{figure*}

\begin{table*}[!htb]
\caption{Comparison of RMSEs of different methods on debutanizer data (test).}
\label{tab:debut_compare}
\centering
\begin{tabular}{@{} lcccccccccc @{}}
\toprule
Label\% & 1\% & 2\% & 5\% & 10\% & 14.2\% & 20\% & 25\% & 33\% & 50\% & 100\% \\
SSVAER     & 0.0764 &	\textbf{0.0522} &	0.0561 	& 0.0498	& \textbf{0.0476}	& \textbf{0.0470} &	\textbf{0.0480} &	\textbf{0.0476} &	\textbf{0.0469} & 0.0516 \\
SVAER     & \textbf{0.0606} &	0.0588 &	0.0541 & \textbf{0.0478} &	0.0507 & 0.0589 &	0.0545	& 0.0490 &	0.0540 &	0.0543\\
SSAE & 0.0822 &	0.0526 & \textbf{0.0494} & 0.0495 & 0.0485 & 0.0529 & 0.0496 & 0.0512 & 0.0521 & \textbf{0.0462} \\
FCNN    & 0.0852 & 0.0541 & 	0.0588 & 0.0517 & 0.0483 & 0.0487 & 0.0487 & 0.0493 & 0.0493 & 0.0495 \\
SSSMM    & NA & NA  & NA & NA & NA & 0.0525 & 0.0731 & 0.0629 & 0.0647 & NA \\
\bottomrule
\end{tabular}
\end{table*}

\begin{table*}[!htb]
\caption{Comparison of RMSEs of different methods on SRU data (test).}
\label{tab:sru_compare}
\centering
\begin{tabular}{@{} lcccccccccc @{}}
\toprule
Label\% & 1\% & 2\% & 5\% & 10\% & 14.2\% & 20\% & 25\% & 33\% & 50\% & 100\% \\
SSVAER     & \textbf{0.0566} &	0.0484 &	\textbf{0.0347} &	0.0322	& 0.0289 &	\textbf{0.0314} &	\textbf{0.0275}	& 0.0285 &	\textbf{0.0268} &	0.0290 \\
SVAER    & 0.0568 &	0.0506 &	0.0354 &	\textbf{0.0322} &	\textbf{0.0275} &	0.0317 &	0.0282 &	0.0320 &	0.0302	& 0.0315 \\
SSAE  & 0.0603 & 0.0499 & 0.0410 & 0.0347 & 0.0308 & 0.0358 & 0.0287 & \textbf{0.0254} & 0.0274 & \textbf{0.0270}\\
FCNN & 0.0653 & \textbf{0.0459} & 0.0461 & 0.0342 & 0.0330 & 0.0362 & 0.0341 & 0.0298 & 0.0367 & 0.0382\\
SSSMM    & NA & NA  & NA & 0.0580 & NA & 0.0351 & 0.0318 & 0.0560 & 0.0445 & NA \\
\bottomrule
\end{tabular}
\end{table*}

In total, there are 10080 process records, the first nine entries are used to construct the input sequence starting at fifth recorded quality variable. We select the last 2071 records as the test dataset, then partition the remaining 8000 records into training and validation data, the training data consists of the first 6000 records, following by the validation data that consists of the rest 2000 records. In terms of determining the number of neurons for the neural network, we follow the inference structure proposed in \cite{sru_struct}, \{20,16,12,6,1\}. For the SVAER and SSVAER, we select \{20,16,12\}, \{12,6,6\}, \{12,6,1\} to be the size of shared encoder, latent encoders, and regressors, respectively.

\section{Results and discussion}
\label{sec:result}

The root mean square errors (RMSE) of the different methods are compared in Tab. \ref{tab:debut_compare}, Tab. \ref{tab:sru_compare} and Fig. \ref{fig:label_trend}. From these tables and graphs, the SSVAER is generally quite robust when the percentage of labelled samples lies between $14.2\%$ and $100\%$. Due to the sizes of data, a comparison test of training with the same number of labelled entries while varying the labelled percentage was not carried out. However, the elbow shape that occurs around 2\% and 10\% on each graph might suggest that in practice, the percentage of labelled entries should not be lower than 10\%. The SSSMM have unfilled entries in the two test cases as it is purely designed for semi-supervised learning, and it has trouble initializing the weights matrices when the percentage of labelled entries is low. 

The SSVAER performs the best in 6 out of 10, and 5 out of 10 in the debutanizer, and SRU datasets, respectively, followed by the SVAER which wins 4 out of 20 in the two datasets. In addition, the SSVAER performs better than the SVAER under supervised learning on these two datasets. From observing Eq. \eqref{eq:label_loss} and Eq. \eqref{eq:label_loss_new}, it can be seen that the benefit is due to the newly added pseudo variation block, and introducing a new DOF to the SVAER could potentially increase its performance on process data. Under supervised learning, SSAE shows the best performances on these two datasets, this is potentially caused by its unique training strategy and the unlabelled entries are only utilized in reconstruction so that it has less regularization effect on the quality variable inference part compared to those in SSVAER and SVAER.

In general, the RMSE increases as the percentage of labelled entries and number of labels in training data decrease. However, none of these methods show a monotonically increasing trend in RMSE when the percentage of labelled entries decreases, and this is likely caused by the differences in sampling frequency for different percentage of labelled entries.

\begin{figure*}[!htb]
\centerline{\includegraphics[width=2.5\columnwidth]{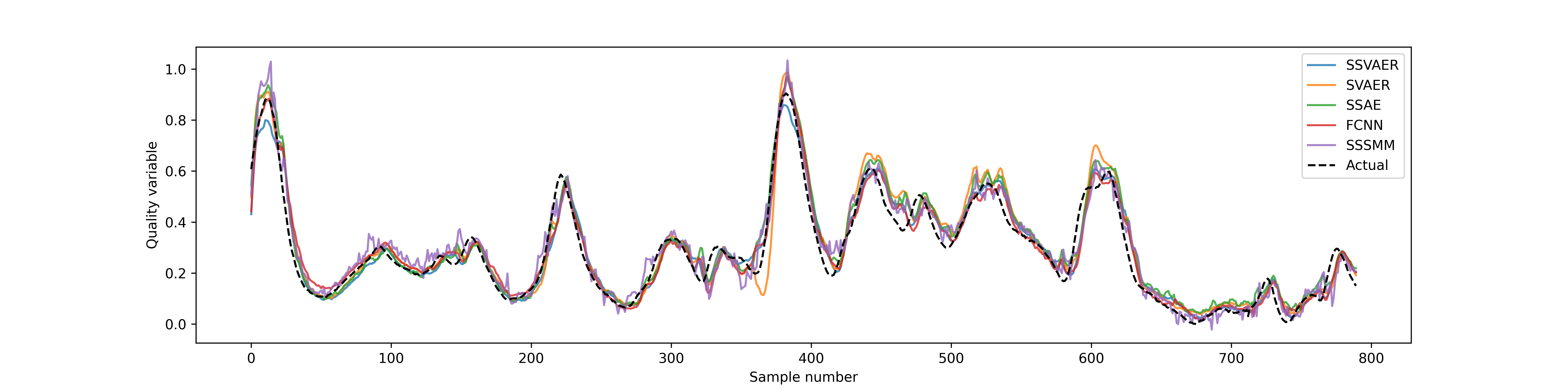}}
\caption{Predictions of each method on debutanizer column (test), trained with 20\% labelled entries.}
\label{fig:debut_plot}
\end{figure*}

\begin{figure*}[!htb]
\centerline{\includegraphics[width=2.5\columnwidth]{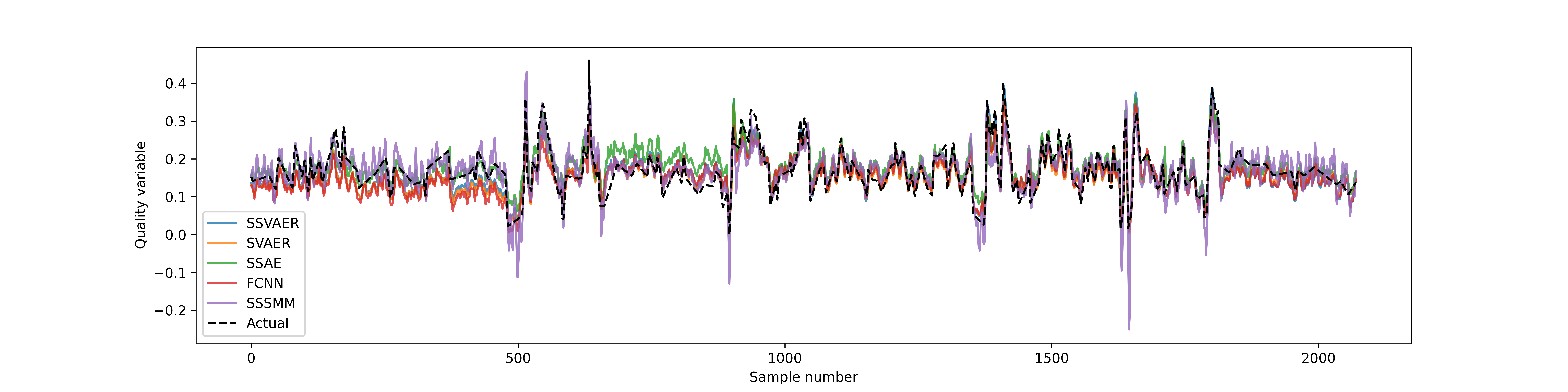}}
\caption{Predictions of each method on SRU (test), trained with 20\% labelled entries.}
\label{fig:sru_plot}
\end{figure*}

In the following part, we will select a labelled percentage of 20\% to demonstrate the result of each method on the two datasets. The prediction results on debutanizer column is shown in Fig. \ref{fig:debut_plot}. Although the partitioning of training and test datasets in our approach is not the same as that in the original SSAE paper \cite{ssae}, the performance of SSAE in our implementation on debutanizer column with 20\% labelled entries, 0.0529 (RMSE), far exceeds their implementation on the same problem with 20\% labelled entries, 0.0801 (RMSE). It shows training strategy, selection of activation function, and weights initialization could have a large impact on the performance of the neural network.

The prediction results on debutanizer column is shown in Fig. \ref{fig:sru_plot}. From the original paper which proposed the \{20,16,12,6,1\} layer structure \cite{sru_struct}, the best result on the SRU test dataset (size of 1000 records) that obtained from picking the least RMSE on test is 0.0279, i.e., the test dataset is used implicitly for training their neural network. It can be seen from Tab. \ref{tab:sru_compare} that all neural network methods can achieve comparable performance on some sampling occasions, and the SSVAER has relatively low variations in RMSE from 100\% to 20\% of labelled entries.

\begin{figure*}[!htb]
\centerline{\includegraphics[width=2.5\columnwidth]{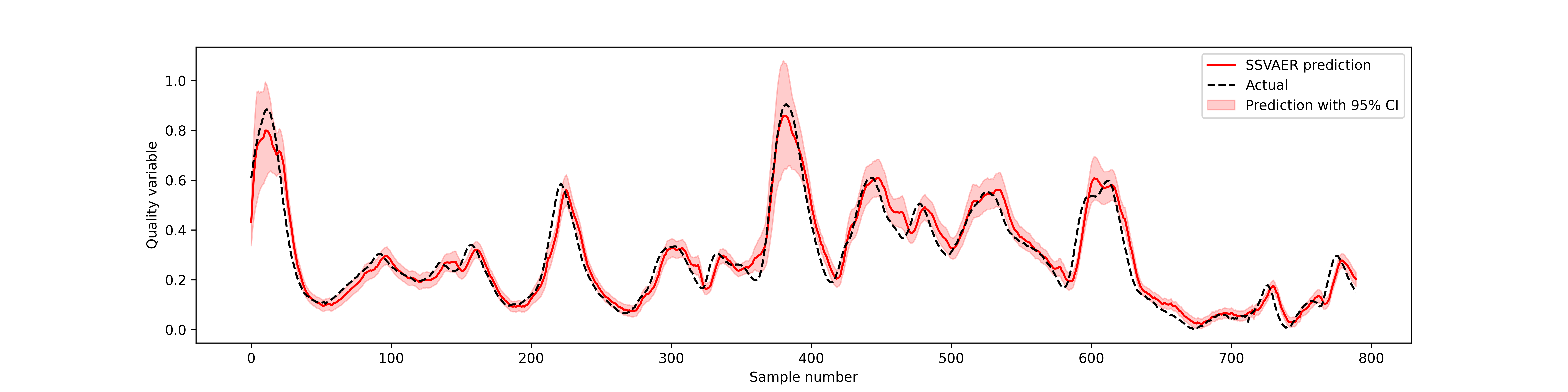}}
\caption{Predictions of SSVAER on debutanizer column (test) with 95\% CI, trained with 20\% labelled entries.}
\label{fig:debut_ci_plot}
\end{figure*}

\begin{figure*}[!htb]
\centerline{\includegraphics[width=2.5\columnwidth]{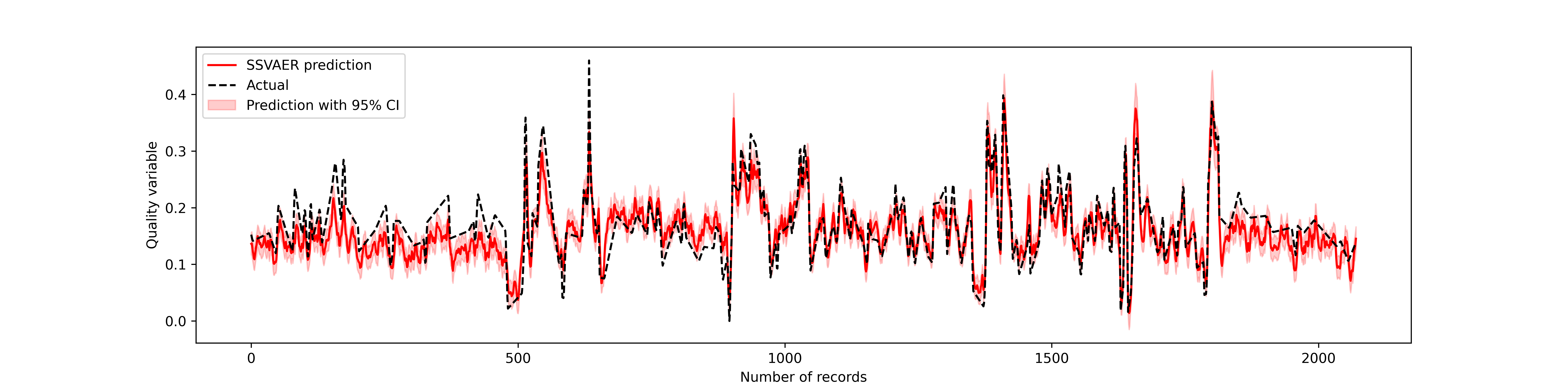}}
\caption{Predictions of SSVAER on SRU (test) with 95\% CI, trained with 20\% labelled entries.}
\label{fig:sru_ci_plot}
\end{figure*}

Since the SSVAER inherits the probabilistic regressor from SVAER, it is also possible to estimate the variance of prediction simultaneously. Taking the sample size as one and assumes the estimated quality variable follows a normal distribution, the 95\% confidence interval (CI) can then be computed. Predictions of quality variables and their CIs are plotted on Fig. \ref{fig:debut_ci_plot} and Fig. \ref{fig:sru_ci_plot}. Although it is difficult to evaluate the variance directly, it can be seen the CI increases when the SSVAER fails to align with the actual records. This feature could be useful when SSVAER/SVAER locates in an ensemble of neural network methods, and the output of the ensemble can be weighted based on the uncertainty of each method.

\begin{figure*}%
    \centering
    \subfloat[\centering Debutanizer column]{{\includegraphics[width=\columnwidth]{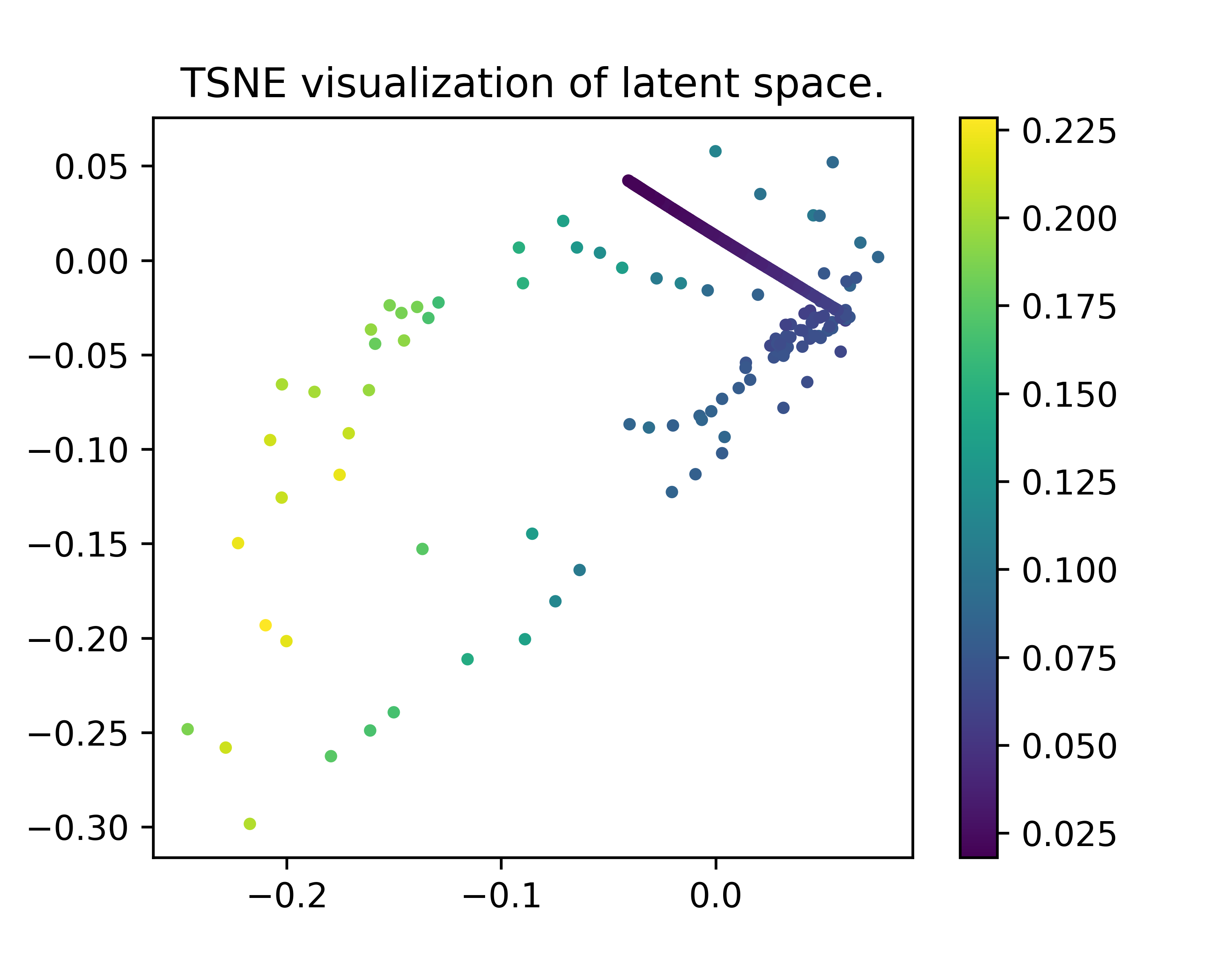} }}%
    \subfloat[\centering SRU]{{\includegraphics[width=\columnwidth]{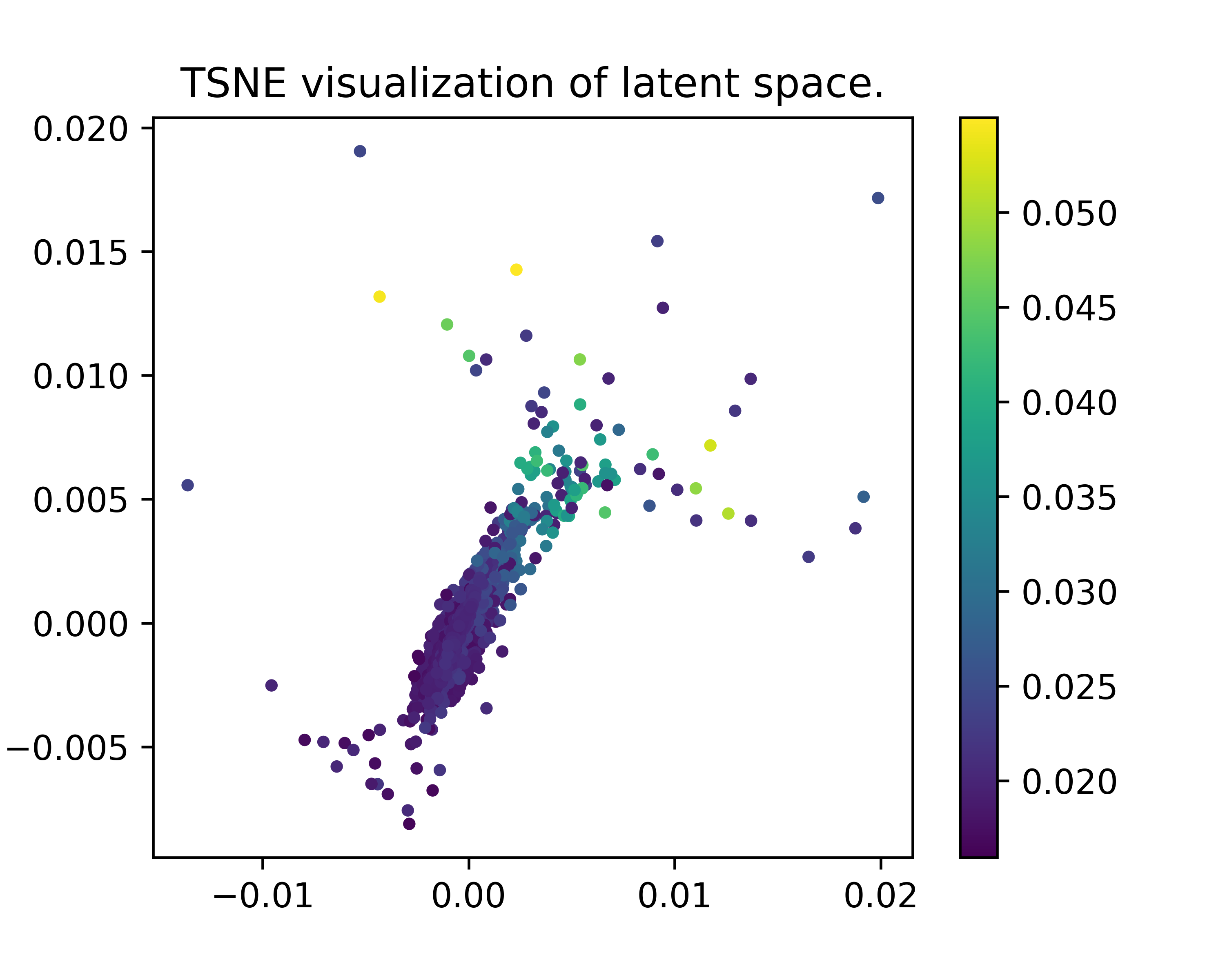} }}%
    \caption{TSNE plots of the latent space of two test datasets (20\% labelled entries). The color bar represents the magnitude of standard deviation.}%
    \label{fig:tsne}%
\end{figure*}

The SVAER can generate a latent space that conditioned on the quality variable \cite{VAER_zhao}, although the latent space of SSVAER is conditioned on both the quality variable and the pseudo variation, it is found to be best described by the variance of the quality variables. As shown in the TSNE plots in Fig. \ref{fig:tsne}, the SSVAER can also obtain a relatively structured latent space. One potential usage with this structured latent space is to apply it on outlier detection, for instance, outliers can be detected either from their distances to clusters or from the corresponding estimated variance.

\section{Conclusion}

In this study, we demonstrate that the SSVAER generally outperforms the SVAER on process data, and it is highly robust when the percentage of labelled entries is in between 14.2\% and 100\%. We showed that the newly introduced regressor block can help the SSVAER to explain more variation in the extracted distribution, and the SSVAER performs the best in 11 out of 20 scenarios, compared to the next best methods which do 4 out of 20. In addition to semi-supervised regression, the probabilistic regressor of the SSVAER also provide an estimation on the variance of the predicted quality variable, and its applications on ensemble learning and outlier detection can be explored further.

The number of neurons and layers of the neural network adopted in this study are based on the structures derived from previous studies, further investigation can be made on determining the correlation between the loss on validation dataset and the accuracy of predictions. Moreover, it is worth exploring how scheduling on KL loss could help SVAER and SSVAER to potentially extract a more meaningful latent distribution, and if it can help on inferencing the quality variable.

\normalsize
\bibliography{main}

\end{document}